\documentclass[a4paper]{jpconf}
\usepackage{graphicx,amsmath,amssymb,subfig}

\newcommand{\beq}{\begin{equation}}
\newcommand{\eq}{\end{equation}}
\newcommand{\req}[1]{(\ref{#1})}

\begin{document}

\title{Modelling conditional probabilities with Riemann-Theta Boltzmann Machines}

\author{Stefano Carrazza$^{1,2}$, Daniel Krefl$^3$, Andrea Papaluca$^1$}

\address{$^1$ TIF Lab, Dipartimento di Fisica, Universit\`a degli Studi di Milano\\
    $^2$ INFN Sezione di Milano, Via Celoria 16, 20133, Milano, Italy\\
    $^3$ Department of Computational Biology, University of Lausanne, Switzerland
}

\ead{stefano.carrazza@unimi.it, daniel.krefl@unil.ch, andrea.papaluca@studenti.unimi.it}

\begin{abstract}
The probability density function for the visible sector of a Riemann-Theta Boltzmann machine can be taken conditional on a subset of the visible units. We derive that the corresponding conditional density function is given by a reparameterization of the Riemann-Theta Boltzmann machine modelling the original probability density function. Therefore the conditional densities can be directly inferred from the Riemann-Theta Boltzmann machine.
\end{abstract}

\section{Introduction}
Modelling the underlying probability density function of a dataset is a non-trivial problem, already in the low dimensional setting. In particular so in the non-normal and multi-modal case. To make things even more complicated, we often do not only want to model the probability density, but as well obtain related quantities like marginalizations, conditionals or the cumulative density function. 

Several techniques to model probability densities of unknown functional form can be found in the literature. To mention a few: Kernel density estimation, mixture models, copulas, normalizing flows, and neural networks. However, each technique comes with its own advantages and drawbacks, and it is fair to say that so far no general use technique is at hand. 

Inspired by Boltzmann machines \cite{HS83}, the authors of \cite{KCHK17} introduced a novel kind of stochastic network, distinguished by an infinite hidden state space given by $\mathbb Z^{N_h}$, with $N_h$ denoting the number of hidden units. Key quantities, like the visible sector probability density function, can be calculated in closed form involving Riemann-Theta functions \cite{M1983}. Therefore, the network has been denoted as Riemann-Theta Boltzmann machine, for short RTBM. In particular, the visible sector density function is given by a novel parametric model, which can be made arbitrarily expressive via increasing the dimension of the hidden state space. The appealing property of this new kind of Boltzmann machine is that the normalization (summation over all states) is given in closed-form in terms of the Riemann-Theta function. The closed form solution allows to keep full analytic control, and in particular to derive related quantities, like for example the corresponding cumulative distribution function or conditional densities. The latter will be discussed in this note.

As conditional distributions are the essential ingredient of Bayes' theorem, modelling conditional distributions has wide applications in machine learning. For instance, in probabilistic modelling a straight-forward application would be as a Bayes classifier.

\subsection{RTBM}

The visible sector probability density function of the Riemann-Theta Boltzmann machine is given by \cite{KCHK17}
\beq\label{PvDef}
P(v)= \sqrt{\frac{\det T}{(2\pi)^{N_v} }}e^{-\frac{1}{2}\left((v + T^{-1} B_v)^t T(v + T^{-1} B_v)\right)} \frac{\tilde\theta(B_h^t + v^t W|Q)}{\tilde\theta(B_h^t - B_v^t T^{-1} W|Q - W^t T^{-1} W)}\,,
\eq
with 
$$
\tilde\theta(z^t|\Omega):=\sum_{n\in \mathbb N^{N_h}} e^{-\frac{1}{2} n^t\Omega n + n^t z }\,,
$$
the Riemann-Theta function. The density $P(v)$ is parameterized by positive definite matrices $Q$ and $T$ of dimension $N_h\times N_h$, respectively $N_v\times N_v$, an arbitrary matrix $W$ of dimension $N_v\times N_h$ and bias vectors $B_h$ and $B_v$ of dimensions $N_h\times 1$, respectively $N_v\times 1$.

As has been discussed for the first time in \cite{KCHK17}, the density $P(v)$ is a powerful density approximator due to its high intrinsic modelling capacity, determined by the number of hidden units $N_h$. For a given set of data samples, the underlying probability density function can be approximated by $P(v)$ via fixing the parameters in a maximum likelihood fashion. However, one should keep in mind that the modelling capacity one can reach in practice is limited by the rather high computational cost of evaluating the Riemann-Theta function, which at present limits $N_h$ to be quite small. For details we refer to \cite{KCHK17} and references therein.
 
It can be shown that $P(v)$ also possesses an interpretation in terms of a specific gaussian mixture model with an infinite number of gaussian constituents \cite{CK18}. In particular, certain useful properties are inherited from the multi-variate gaussian density, like for example functional invariance under affine transformations of the datapoints $v$. 

In this note we will discuss another useful property, which one may see as well to be inherited from the multi-variate gaussian. Namely, that the conditional density functions can again be expressed in terms of the original density $P(v)$, albeit under a different parameterization. Therefore, once we learned an approximation $P(v)$ of a multi-variate density via a RTBM, we obtain all the conditional densities automatically, as we will show in the following section.

\section{Conditional probability}

\subsection{Derivation}\label{deriv}

We take $v = (y_1,\dots,y_m,d_1,\dots,d_n)$ with $m+n=N_v$ and consider the conditional density function
$$
P(y|d) = \frac{P(v)}{P(d)}\,,
$$
with $P(v)$ given by the density \req{PvDef} and $P(d)$ its marginalization
\beq\label{PdDef}
P(d) = \int_{-\infty}^\infty [dy]\, P(y,d)\,.
\eq
It is usefull to decompose the parameter matrices $T,W$ and $B_v$ of the density \req{PvDef} into the following block forms:
$$
T=\left(\begin{array}{@{}c|c@{}}
  \begin{matrix}
  \bar{T}_0
  \end{matrix}
  & \quad \bar{T}_1^t \quad \quad \\
\hline \\
  \quad \bar{T}_1 \quad &
  \begin{matrix}
  \tilde{T}
  \end{matrix}\\
  \\
\end{array}\right)\,,
$$
with $\bar{T}_0\,$ a $m\times m$ square matrix, $\bar{T}_1\,$ a $n\times m$ rectangular matrix and $\tilde T\,$ a $n\times n$ square matrix.\\
Similarly,
$$
W=\left(\begin{array}{cc}
\quad \quad W_0 \quad \quad \\
\hline \\
\quad \quad W_1 \quad \quad\\
\\
\end{array}
\right)\,,\,\,\,\,\,
B_v=\left(\begin{array}{cc}
B_{v,0}\\
\hline
 B_{v,1}\\
\end{array}
\right)\,,
$$
with $W_0$ and $W_1$ rectangular matrices of dimension $m\times N_h$, respectively $n\times N_h$. $B_{v,0}$ and $B_{v,1}$  are column vectors of size $m$, respectively, $n$.

Imposing the above block structure onto the terms $v^t T v$, $B_v^tv$ and $v^tW$ occurring in the density \req{PvDef} gives
\beq
\begin{split}
  v^t T v&=y^t\bar{T}_0y+2d^t\bar{T}_1y+d^t\tilde{T}d\,,\\
   B_v^tv&= B_{v,0}^ty+B_{v,1}^td\,,\\
   v^tW&=y^tW_0+d^tW_1\,,\\
\end{split}
\eq
and leads to the following expression for the joint density $P(v) = P(y,d)$:
\begin{multline}\label{P(y,d)}
  P(y,d)=\sqrt{\frac{\det T}{(2\pi)^{Nv}}}\quad e^{-\frac{1}{2}y^t\bar{T}_0y-d^t\bar{T}_1y-\frac{1}{2}d^t\tilde{T}d-B_{v,0}^ty-B_{v,1}^td-\frac{1}{2}B^t_vT^{-1}B_v}\\\times\frac{\tilde\theta(B^t_h+y^tW_0+d^tW_1|Q)}{\tilde\theta(B^t_h-B^t_vT^{-1}W|Q-W^tT^{-1}W)}\,.
\end{multline}
Note that the terms $B_v^tT^{-1}B_v$, $B_v^tT^{-1}W$ and $W^tT^{-1}W$ are not written in factorized form because they do not include $y$ and therefore can be pulled out of the marginalization integration \req{PdDef}.  After explicitation of the theta function the integral we have to solve to obtain $P(d)$ reads
$$I=\int_{-\infty}^{+\infty} [dy] \quad e^{-\frac{1}{2}y^t\bar{T}_0y\,+\,(n^tW_0^t-d^t\bar{T}_1-B_{v,0}^t)\,y}\,,$$
which is the well known generalized gaussian integral
$$\int_{-\infty}^\infty [dx] \quad e^{-\frac{1}{2}x^tA x + b^tx}=\sqrt{\frac{(2\pi)^n}{\det A}}\quad e^{\frac{1}{2}b^tA^{-1} b}.$$

Hence, we find the explicit expression
\begin{multline}\label{P(d)}
  P(d)=\sqrt{\frac{\det T}{(2\pi)^{Nv}}}\quad e^{-\frac{1}{2}d^t\tilde{T}d-B_{v,1}^td-\frac{1}{2}B^t_vT^{-1}B_v}\sqrt{\frac{(2\pi)^m}{\det \bar T_0}}\quad e^{(B_{v,0}+\bar T_1^td)^t\bar T_0^{-1}(B_{v,0}+\bar T_1^td)}\\\times\frac{\tilde\theta(B_h^t+d^tW_1-(B_{v,0}+\bar T_1^td)^t\bar T_0^{-1}W_0|Q-W_0^t\bar T_0^{-1}W_0)}{\tilde\theta(B^t_h-B^t_vT^{-1}W|Q-W^tT^{-1}W)}\quad ,
  \end{multline}
which in turn leads to the desired conditional probability
\begin{multline}
  P(y|d)=\frac{P(v)}{P(d)}=\sqrt{\frac{\det \bar{T}_0}{(2\pi)^{m}}}\quad e^{-\frac{1}{2}y^t\bar{T}_0y-(B_{v,0}+\bar{T}_1^td)^ty-\frac{1}{2}(B_{v,0}+\bar{T}_1^td)^t\bar{T}_o^{-1}(B_{v,0}+\bar{T}_1^td)}\,\\ \times\frac{\tilde\theta(B^t_h+d^tW_1+y^tW_0|Q)}{\tilde\theta(B^t_h+d^tW_1-(B_{v,0}+\bar{T}_1^td)^t\bar{T}_0^{-1}W_0|Q-W_0^t\bar{T}_0^{-1}W_0)}\,.
\end{multline}

\begin{figure}
  \begin{center}
    \includegraphics[scale=0.22]{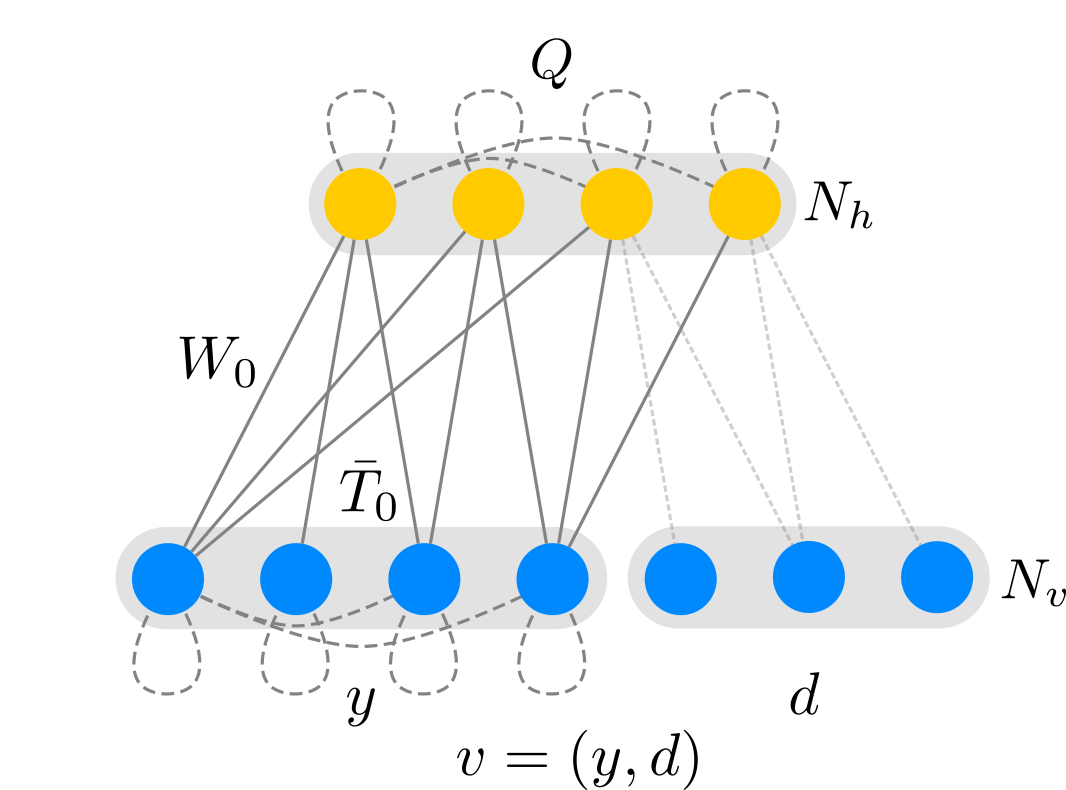}
    \caption{\footnotesize{Graphical representation of a conditional
        probability RTBM architecture.}}\label{ConditionalNet}
    \label{architecture}
  \end{center}  
\end{figure}

Note that $P(y|d)$ as given above is easily seen to correspond to a probability density function $P(v)$ of a $N_v=m$ RTBM with the following reparameterization:
\beq\label{Ptransform}
\begin{split}
T &\rightarrow \bar{T}_0\,,\\
W &\rightarrow W_0\,,\\
B_v &\rightarrow B_{v,0}+\bar{T}_1^td\,,\\
B_h &\rightarrow B_h+W_1^td\,.
\end{split}
\eq
We conclude that starting from a ``parent'' RTBM modelling a multidimensional density, we can generate ``child'' RTBMs modelling its conditional probabilities simply by choosing the parameters accordingly. For illustration, we sketched the surviving parameters of the corresponding network architecture in figure \ref{ConditionalNet}.

\subsection{Examples}

\begin{figure}
  \begin{center}
    \includegraphics[scale=0.35]{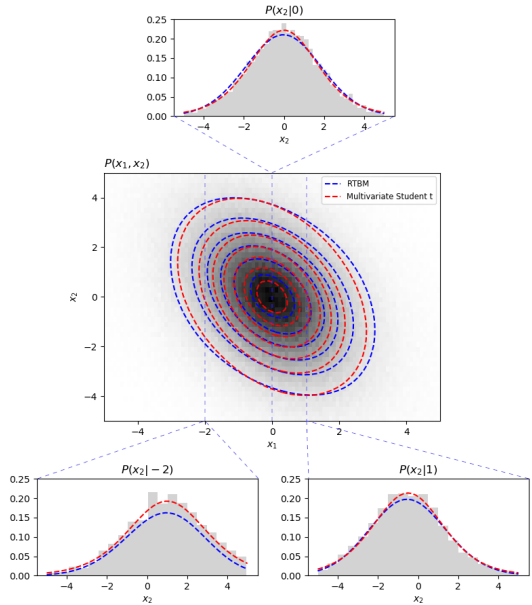}
    \caption{\footnotesize{In the center the two dimensional histogram sampled by a multivariate Student $t$-distribution \req{t_distr} with parameters \req{t_par} is shown. The contour plot of the trained RTBM \req{t_ex_par} is shown in blue, the analytic distribution in red. On the bottom the comparison between the analytic conditionals given by \req{t_cond} with the ones generated using \req{Ptransform} are shown.}}\label{student}
  \end{center}  
\end{figure}

As an example of what we have discussed above, let us consider the multivariate Student's $t$-distribution:
\beq\label{t_distr}
f(x)=\frac{\Gamma((v+p)/2)}{\Gamma(v/2)\,(v\pi)^{\frac{p}{2}}\,|\Sigma|^{\frac{1}{2}}}\bigg[1+\frac{1}{v}\,(x-\mu)^t\Sigma^{-1}(x-\mu)\bigg]^{-\frac{v+p}{2}}\,.
\eq
As derived in \cite{student_cond1}, \cite{student_cond2}, it possesses an analytic expression for its conditional. For $x=(x_1,x_2)$ one finds that
\beq\label{t_cond}
x_2\,|\,x_1\,\sim\,t_{p_2}\bigg(\mu_{2|1},\frac{v+d_1}{v+p_1}\,\Sigma_{22|1},v+p_1\bigg)\,.
\eq
Note that the conditional is again a $t$-distribution. This allows us to easily compare the analytical conditional with the one obtained from an RTBM trained to fit the corresponding $t$-distribution.

We consider a $t$-distribution with the following parameters:
\beq\label{t_par}
  \mu=(0,0),\quad
  \Sigma=\left(
  \begin{matrix}
   2  & -1  \\
   -1 & 4  \\
  \end{matrix}
  \right),\quad
  v=6,
  \eq
A RTBM with $N_v=2$ and $N_h=2$ is trained with the CMA-ES algorithm on $5000$ samples thereof. The best solution was found at a log-likelihood loss of $\sim 1.3\cdot 10^4$. The found RTBM parameters fitting the above $t$-distribution read:
  \beq\label{t_ex_par}
\begin{split}
W& = \left(
\begin{matrix}
-1.11 & 1.02 \\
-0.66 & 0.60\\
\end{matrix}
\right)\,,\,\,\,\,\,
T = \left(
\begin{matrix}
0.56 &  0.18 \\
0.18 & 0.30\\
\end{matrix}
\right)\,,\,\,\,\,\,
B_v = \left(
\begin{matrix}
0  \\
0  \\
\end{matrix}
\right)\,,\,\,\,\,\,\\
B_h&=\left(
\begin{matrix}
  8.22 \\
  17.40\\
\end{matrix}
\right)\,,\,\,\,\,\, 
Q =\left(
\begin{matrix}
  24.15 & -0.44\\
  -0.44 & 41.57
\end{matrix}
\right)\,.
\end{split}
\eq
The analytic contribution and its RTBM based fit are shown in figure \ref{student}. The figure also shows three examples of conditionals derived following section \ref{deriv} and the corresponding analytic solution obtained from \req{t_cond}.

In order to quantify the error in modelling the conditionals with the RTBM, we calculated the mean squared error (MSE) between the analytic distribution and the RTBM fit at the sample points used for training. The results are shown in table \ref{mse}.

\begin{table}
  {\footnotesize
\begin{center}
  \begin{tabular}{||c|c||c|c||c|c||}
    \hline
    \multicolumn{2} {||c||} {Multivariate Student $t$} &
    \multicolumn{2} {c||} {2D Example} &
    \multicolumn{2} {c||} {3D Example} \\
    \hline
        {Conditional} & {MSE} & {Conditional} & {MSE} & {Conditional} & {MSE} \\
        \hline
            {$P(x_2|-2)$} & {$3.255\cdot10^{-4}$} & {$P(y|2)$} & {$1.176\cdot10^{-4}$} & {$P(y_1,y_2|-0.4)$} & {$4.953\cdot10^{-5}$}\\
            \hline
                {$P(x_2|0)$} & {$4.083\cdot10^{-5}$} & {$P(y|1.3)$} & {$6.888\cdot10^{-5}$} & {$P(y_1,y_2|-0.6)$} & {$6.775\cdot10^{-5}$}\\
                \hline
                    {$P(x_2|1)$} & {$4.433\cdot10^{-5}$} & {$P(y|0.4)$} & {$2.538\cdot10^{-4}$} & {$P(y_1,y_2|-0.8)$} & {$5.304\cdot10^{-5}$}\\
                    \hline
  \end{tabular}
\caption{\footnotesize{MSE calculated for each of the conditional distributions shown in figure \ref{student}, \ref{2dcond}, \ref{3dcond}. Note that for the 2D and 3D examples, the MSE is calculated with respect to the empirical conditional derived from the relative histogram. }} \label{mse}
\end{center}
}
\end{table}

For illustration purposes, we consider two further examples for the
conditional densities obtainable through relation
\req{Ptransform}. We only show results with $N_v=2$ and $N_v=3$ in
order to simplify the visualization of results. However one should note that the current
methodology is valid for higher dimensional examples as well.

On both examples the RTBM parameters are
initialized manually in order to achieve a more complicated distribution $P(v)$ than the $t$-distribution discussed above. The conditional distributions are obtained as before
following the transformations presented in section \ref{deriv}. On
both examples we compare the resulting conditional
probabilities with the empirical distributions obtained using the RTBM
sampling algorithm presented in \cite{CK18}.

\begin{figure}
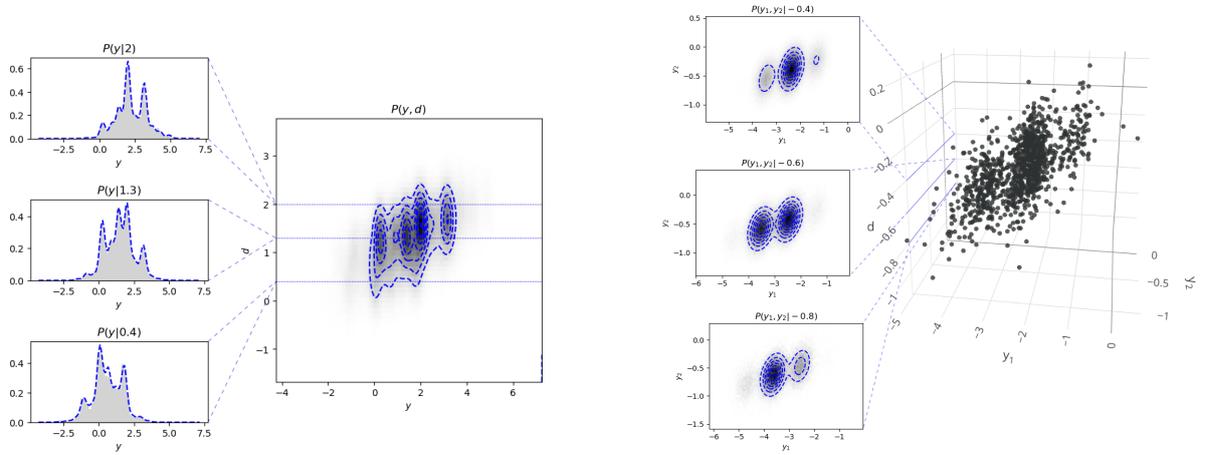

  \centering \subfloat[\footnotesize{On the right, the two dimensional
      histogram of a RTBM with $N_h=4$, $N_v=2$ and initialized with
      parameters \req{2dExampleRTBM} is shown. The contour plot is
      represented in blue. On the left the conditional probabilities
      $P(y|d)$ for three different values $d=2,\,1.3,\,0.4$ are
      shown. The relative one dimensional histogram is shown in
      grey.}]{
    \includegraphics[width=0.45\textwidth]{fig/distr3_y_slices-1.png}
      \label{2dcond}}
  \hfill
  \subfloat[\footnotesize{On the right, a sampling of the three
      dimensional distribution given by a RTBM with $N_v=3$, $N_h=1$
      and parameters \req{3dExampleRTBM}. On the left, the contour
      plots of the conditional probabilities $P(y_1,y_2|d)$,
      represented by the dashed blue curves, evaluated at
      $d=-0.4,\,-0.6,\,-0.8$ are shown respectively. The corresponding
      two dimensional empirical histograms are shown in
      gray.}]{
    \includegraphics[width=0.45\textwidth]{fig/3dconditional-edited.png}
    \label{3dcond}}
  \caption{\footnotesize{Examples of conditional probabilities.}}
\end{figure}

The RTBM in the two dimensional example with $N_v=2$ and $N_h=4$ is defined via the following parameter choice:
\beq\label{2dExampleRTBM}
\begin{split}
  W& = \left(
\begin{matrix}
18.54 & 3.02 & -12.89 & -5.45 \\
0.46 & 1.01 & -1.32 & -5.54 \\
\end{matrix}
\right)\,,\,\,\,\,\,
T = \left(
\begin{matrix}
28.77 & 0 \\
0 & 6.3 \\
\end{matrix}
\right)\,,\,\,\,\,\,
B_v = \left(
\begin{matrix}
-1.76  \\
-2.69  \\
\end{matrix}
\right)\,,\\
B_h&=\left(
\begin{matrix}
-0.31  \\
2.29  \\
1.65 \\
-2.73\\
\end{matrix}
\right)\,,\,\,\,\,\,
Q =\left(
\begin{matrix}
15.48 & 8.82 & -3.19 & -3.67 \\
8.82 & 17.99 & 8.94 & -4.04 \\
-3.19 & 8.94 & 15.74 & 4.14 \\
-3.67 & -4.04 & 4.14 & -5.54\\
\end{matrix}
\right)\,.
\end{split}
\eq
The corresponding distribution and derived conditional distributions are shown in figure \ref{2dcond}. 

For the three dimensional example, we define a RTBM with $N_v=3$ and $N_h=1$. The chosen parameters are
\beq\label{3dExampleRTBM}
\begin{split}
W& = \left(
\begin{matrix}
-15.76 \\
2.29\\
2.09 \\
\end{matrix}
\right)\,,\,\,\,\,\,
T = \left(
\begin{matrix}
16.02 & -6.52 & -6.76 \\
-6.52 & 29.04 & -2.56\\
-6.76 & -2.56 & 42.16\\
\end{matrix}
\right)\,,\,\,\,\,\,
B_v = \left(
\begin{matrix}
1.08  \\
-0.67  \\
4.86 \\
\end{matrix}
\right)\,,\,\,\,\,\,\\
B_h&=\left(
\begin{matrix}
3.17 \\
\end{matrix}
\right)\,,\,\,\,\,\, 
Q =\left(
\begin{matrix}
19.18\\
\end{matrix}
\right)\,.
\end{split}
\eq
The corresponding distribution and examples of obtained two dimensional conditional densities are show in figure \ref{3dcond}.

The MSE between the conditional distributions derived from the RTBM and the empirical conditionals derived from histograms are listed in table \ref{mse}. However, one should keep in mind that the purpose of the MSE calculation in these two examples is soley to illustrate that the relation \req{Ptransform} is correct. By construction, the RTBM constitute the true (analytic) underlying distribution for these examples.

\section*{Acknowledgments}
S.C. is supported by the European Research Council under the European
Union’s Horizon 2020 research and innovation Programme (grant
agreement number 740006).

\end{document}